\documentclass{article} 
\usepackage{iclr2020_conference,times}

\usepackage{amsmath,amsfonts,bm}

\def\eqref#1{equation~\ref{#1}}

\def\1{\bm{1}}

\DeclareMathAlphabet{\mathsfit}{\encodingdefault}{\sfdefault}{m}{sl}
\SetMathAlphabet{\mathsfit}{bold}{\encodingdefault}{\sfdefault}{bx}{n}

\usepackage{url}

\usepackage{biblatex}
\title{Neural Machine Translation for South Africa's Official Languages}

\author{Laura Martinus, Jason Webster, Joanne Moonsamy, Moses Shaba Jnr, Ridha Moosa, \\ \textbf{\& Robert Fairon}  \\
EXPLORE Data Science Academy, South Africa\\
\texttt{\{laura,jason,joanne,moses,ridha,robert\}@explore-ai.net}
}

\iclrfinalcopy 
\begin{document}

\maketitle

\begin{abstract}
Recent advances in neural machine translation (NMT) have led to state-of-the-art results for many European-based translation tasks. However, despite these advances, there is has been little focus in applying these methods to African languages. In this paper, we seek to address this gap by creating an NMT benchmark BLEU score between English and the ten remaining official languages in South Africa.
\end{abstract}

\section{Introduction}
Attention-based neural networks, such as Transformer \cite{transformer}, and pre-training methods, such as BERT \cite{bert}, are established state-of-the-art methods for sequence modelling and translation tasks. However, these methods and their applications typically focus on European languages, with little focus towards African languages. Community driven efforts such as Masakhane\footnote{\url{www.masakhane.io}} have since been established, and have made great strides towards benchmarking neural machine translation (NMT) tasks for all African languages.

In this paper, we provide a benchmark BLEU score for translation tasks between English and the ten remaining official South African languages. We train ten Transformer models on the JW300 parallel corpus \cite{jw300} and provide BLEU scores on a derived testing subset. We then compare our model to a previously generated benchmark \cite{abbott2019benchmarking} using the Autshumato evaluation set \cite{mckellar2020dataset}.

\section{Languages}
The 11 official languages of South Africa are categorised into five language family groups. The largest of these groups is the conjunctively written Nguni languages which include siSwati, isiZulu, isiXhosa and isiNdebele. These languages are spoken widely through the eastern parts of South Africa by roughly 45 \% of the population \cite{statssa_2016}. 

The Sotho-Tswana branch includes the Sesotho, Northern Sotho, and Setswana \cite{hickey2019english}. This disjunctively written family of languages is spoken predominantly through the central parts of South Africa as well as in the neighbouring countries of Lesotho and Botswana. Northern Sotho and Setswana are closely related and mutually intelligible. Tshivenda and Xitsonga fall into their own categories, and their native speakers are concentrated in the Limpopo and Mpumalanga provinces of South Africa respectively \cite{eberhard_simons_fenning}. 

The final group consists of the Germanic languages English and Afrikaans. Afrikaans is a West-Germanic language, analytic in nature, and originated from Dutch settlers in the 18th century \cite{roberge2002afrikaans}.

\section{Model Architecture}

Models were created using Joey NMT \cite{joey2019}, a minimalistic NMT toolkit created for educational purposes. In all of our models, we used 6 layers with 4 attentions heads in both the encoder and decoder, with an embedding dimension of 256 and a hidden dimension of 256. Each model was trained on an Nvidia Tesla P-100 GPU for approximately 6 hours, or until convergence was reached.

\section{Data}

Our models are trained using the JW300 parallel corpus \cite{jw300}, where a subset of this data is set aside for testing\footnote{Test sets can be found at \url{https://github.com/masakhane-io/masakhane/tree/master/jw300_utils/test}}. The number of parallel sentences for each language is given in Table~\ref{jw300-parallel}. We remove any duplicate sentences, and use \texttt{fuzzywuzzy}\footnote{\url{https://github.com/seatgeek/fuzzywuzzy}} to remove sentences with a similarity score over 0.95. Finally, we apply byte-pair encoding \cite{bpe} with 40~000 codes to each combined source-target corpus.

We also make use of the Autshumato evaluation set \cite{mckellar2020dataset} for testing purposes only. Each set is made up of 500 parallel sentences for all 11 official languages of South Africa, each of which were translated by four professional translators.

\begin{table}[t]
    \begin{minipage}[b]{.45\linewidth}
        \caption{Number of English-Language parallel sentences in the JW300 dataset}
        \vspace{1em}
        \centering
        \begin{tabular}{l|cc}
        \label{jw300-parallel}
        &\multicolumn{2}{c}{\bf Parallel sentences} \\
        \bf Language & \bf Train set & \bf Test Set \\
        \hline\hline
        Afrikaans      &   995 740 & 2 682\\
        isiNdebele     &    53 640 & 2 615\\
        isiXhosa       &   786 371 & 2 711\\
        isiZulu        &   956 474 & 2 711\\
        Northern Sotho &    98 668 & 2 711\\
        Sesotho        &   831 230 & 2 672\\
        Setswana       &   791 068 & 2 714\\
        siSwati        &    98 668 & 2 681\\
        Tshivenda      &   204 153 & 2 719\\
        Xitsonga       &   766 752 & 2 715\\
        \end{tabular}
    \end{minipage}%
    \quad
    \begin{minipage}[b]{.45\linewidth}
        \caption{English-Language BLEU scores for the Autshumato and JW300 test sets}
        \vspace{1em}
        \centering
        \begin{tabular}{l|cc|c}
        \label{table:results}
        & \multicolumn{2}{|c|}{\bf Autshumato} & \bf JW300 \\
        \bf Language  &\bf From \cite{abbott2019benchmarking} & \bf Ours
        &\bf Ours
        \\ \hline\hline
        Afrikaans      & 20.60 & 16.67 & 58.47 \\
        isiNdebele     &   -   &  0.00 & 23.58\\
        isiXhosa       &   -   &  1.42 & 37.11\\
        isiZulu        &  1.34 &  1.56 & 44.07\\
        Northern Sotho & 10.94 &  8.26 & 45.95\\
        Sesotho        &   -   &  5.21 & 41.23\\
        Setswana       & 15.60 &  8.13 & 46.91\\
        siSwati        &   -   &  0.41 & 35.75\\
        Tshivenda      &   -   &  5.15 & 52.27\\
        Xitsonga       & 17.98 &  7.28 & 46.41\\
        \end{tabular}
    \end{minipage}
\end{table}

\section{Results}

Table~\ref{table:results} provides a comparison of the five benchmarked models from \cite{abbott2019benchmarking} and our models. Most of our models receive a BLEU score of over 40, with exception of isiNdebele, Northern Sotho, and siSwati -- corresponding to our three smallest data sets with less than 100~000 parallel sentences.

A possible explanation for the low Autshumato scores as compared to \cite{abbott2019benchmarking} may be that the Autshumato domain is governmental, while the JW300 training set domain is religious. The models in \cite{abbott2019benchmarking} were trained with the Autshumato training data, so their domain was closer to the testing data than our models. Another explanation is that the Autshumato evaluation set proposes many hypothetical translations from the same English source sentence. Our model only offers a single translation for a source sentence, so even if our model proposes a human level translation, we may receive a low BLEU score as it would not match the other human-level translations.

\section{Conclusion and Future Work}

We have presented a benchmark NMT score between English and all remaining official South African languages using the JW300 dataset for training and testing. We also tested our models on the Autshumato evaluation set and found that our models do not perform as well as models that were trained within the domain of the Autshumato training data. This low performance could be the result of two cases: our models failing to generalise outside of its training domain or; the fact that the Autshumato test set proposes many hypothetical translations for a single English source sentence, leading to a low BLEU score. Future work shall investigate both cases.


\newpage
\printbibliography


\end{document}